\documentclass[11pt,a4paper]{article}
\usepackage[hyperref]{naaclhlt2019}
\usepackage{times}
\usepackage{latexsym}

\usepackage{url}
\usepackage{graphicx}  
\usepackage{amsfonts}
\aclfinalcopy 
\def\aclpaperid{443} 

\title{Semantic Role Labeling with Associated Memory Network}

\author{
  Chaoyu Guan$^\dag$, Yuhao Cheng$^\dag$, Hai Zhao$^\dag$\thanks{Corresponding author. This paper was partially supported by 
  National Key Research and Development Program of China (No. 2017YFB0304100), 
  National Natural Science Foundation of China (No. U1836222 and No. 61733011) and
  Key Project of National Society Science Foundation of China (No. 15-ZDA041).} \\
  $^\dag$Department of Computer Science and Engineering, Shanghai Jiao Tong University\\
  $^\dag$Key Laboratory of Shanghai Education Commission for Intelligent Interaction \\
  and Cognitive Engineering, Shanghai Jiao Tong University, Shanghai, China \\
  {\tt \{351549709,cyh958859352\}@sjtu.edu.cn} \\
  {\tt zhaohai@cs.sjtu.edu.cn} \\}

\date{}

\begin{document}
\maketitle
\begin{abstract}
Semantic role labeling (SRL) is a task to recognize all the predicate-argument pairs of a sentence, which has been in a performance improvement bottleneck after a series of latest works were presented. This paper proposes a novel syntax-agnostic SRL model enhanced by the proposed associated memory network (AMN), which makes use of inter-sentence attention of label-known associated sentences as a kind of memory to further enhance dependency-based SRL. In detail, we use sentences and their labels from train dataset as an associated memory cue to help label the target sentence. Furthermore, we compare several associated sentences selecting strategies and label merging methods in AMN to find and utilize the label of associated sentences while attending them. By leveraging the attentive memory from known training data, Our full model reaches state-of-the-art on CoNLL-2009 benchmark datasets for syntax-agnostic setting, showing a new effective research line of SRL enhancement other than exploiting external resources such as well pre-trained language models.
\end{abstract}

\section{Introduction}

Semantic role labeling (SRL) is a task to recognize all the predicate-argument pairs of a given sentence and its predicates. It is a shallow semantic parsing task, which has been widely used in a series of natural language processing (NLP) tasks, such as information extraction \cite{liu2016learning} and question answering \cite{abujabal2017automated}.

Generally, SRL is decomposed into four classification subtasks in pipeline systems, consisting of predicate identification, predicate disambiguation, argument identification, and argument classification. In recent years, great attention \cite{zhou2015end,marcheggiani2017simple,he2017deep,he2018jointly,he2018syntax} has been turned to deep learning method, especially Long Short-term Memory (LSTM) network for learning with automatically extracted features. \cite{zhou2015end} proposed the first end-to-end recurrent neural network (RNN) to solve the SRL task. \cite{marcheggiani2017simple} studied several predicate-specified embedding and decoding methods. \cite{he2017deep} delivered a full study on the influence of RNN training and decoding strategies. Whether to use the syntactic information for SRL is also studied actively \cite{he2017deep,he2018syntax}.

Since the recent work of \cite{marcheggiani2017simple}, which surprisingly shows syntax-agnostic dependency SRL for the first time can be rival of syntax-aware models, SRL has been more and more formulized into standard sequence labeling task on a basis of keeping syntax unavailable. A series of work on SRL received further performance improvement following this line through further refining neural model design \cite{he2018jointly}. Different from all previous work, we propose to introduce an associated memory network which builds memory from known data through the inter-sentence attention to enhance syntax-agnostic model even further.

Inspired by the observation that people always refer to other similar problems and their solutions when dealing with a problem they have never seen, like query in their memory, we want to utilize similar known samples which include the associated sentences and their annotated labels to help model label target sentence. To reach such a goal, we adopt a memory network component,
and use inter-sentence attention to fully exploit the information in memory.


Based on Memory Network \cite{Weston2015Memory,sukhbaatar2015end}, \cite{miller2016key} proposed Key-Value Memory Network (KV-MemNN) to solve Question Answering problem and gain large progress. Our proposed method is similar to KV-MemNN, but with a different definition of key-value and different information distilling process. Thus, we propose a carefully designed inter-sentence attention mechanism to handle it.

Recently, there are also some attempts to make use of attention mechanism in SRL task. \cite{Tan2017Deep,strubell2018linguistically} focus on self-attention, which only uses the information of the input sentence as the source of attention. \cite{cai2018full} makes use of biaffine attention \cite{DBLP:journals/corr/DozatM16} for decoding in SRL, which was the current state-of-the-art (SOTA) in CoNLL-2009 benchmark as this work was embarking. Different from all previous work, we utilize inter-sentence attention to help model leverage associated information from other known sentences in the memory.

To our best knowledge, this is the first time to use memory network in the SRL task. Our evaluation on CoNLL-2009 benchmarks shows that our model outperforms or reaches other syntax-agnostic models on English, and achieves competitive results on Chinese, which indicates that memory network learning from known data is indeed helpful to SRL task.

There are several SRL annotation conventions, such as PropBank \cite{bonial2012english} and FrameNet \cite{baker1998berkeley}. This paper focuses on the former convention. Under PropBank convention, there are two role representation forms, which are span-based SRL, such as CoNLL 2005 and CoNLL 2012 shared tasks, and dependency-based SRL, such as CoNLL 2009 shared task. The former uses span to represent argument, while the latter uses the headword of the span to represent the argument. As the latter has been more actively studied due to dependency style SRL for convenient machine learning, we will focus on dependency SRL only in this work.

Given a sentence $\mathbf{S}$, the goal of dependency SRL task is to find all the predicate-argument pairs $(p,a)$. The following shows an example sentence with semantic role labels marked in subscripts.

\normalsize \textbf{She}$_{\scriptsize A0}$ \normalsize has \normalsize\textbf{lost}$_{\scriptsize v}$ \textbf{it}$_{\scriptsize A1 }$ \normalsize \textbf{just}$_{\scriptsize ARGM-MNR}$

as quickly.

Here, \emph{v} means the predicate, \emph{A0} means the agent, \emph{A1} means the patient and \emph{ARGM-MNR} means how an action \emph{v} is performed.

In the rest of this paper, we will describe our model in Section \ref{model}. Then, the experiment set-up and results are given in Section \ref{exp}. Related works about SRL and attention mechanism will be given in Section \ref{related}. Conclusions and future work are drawn in Section \ref{future}.

\section{Model \label{model}}
\begin{figure*}[htbp]
	\centering\includegraphics[width=\textwidth]{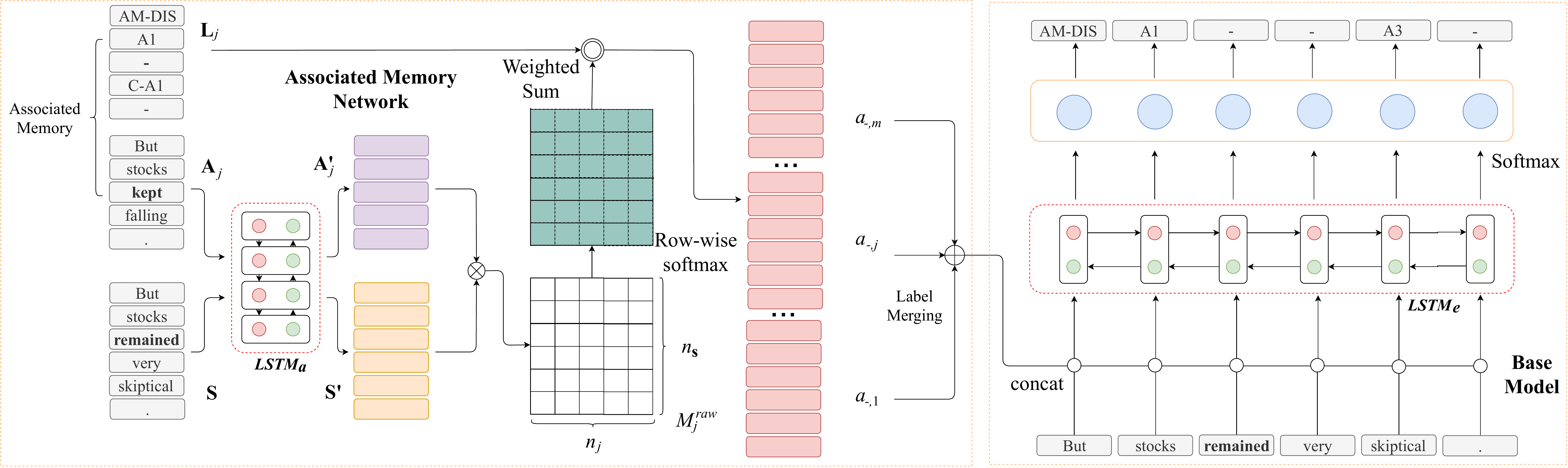}
	\caption{Semantic role labeling with associated memory network, where ${\rm S}$ is the input sentence with its length $n_{\rm S}$. ${\rm A}_j$ is the $j^{th}$ associated sentence of ${\rm S}$ with its label sequence ${\rm L}_j$ and its length $n_j$. ${\rm S}'$ and ${\rm A}'_j$ are the result of $LSTM_1$ with ${\rm S}$ and ${\rm A}_j$ as input respectively. $d_{ae}$ is the dimension of argument embedding. $M^{raw}_j$ is the raw attention matrix of ${\rm A}_j$ and ${\rm S}$. $a_{-,j} = [a_{1,j},a_{2,j},...,a_{n_{\rm S},j}]$ is the associated-sentence-specified attention embedding.}\label{2}
\end{figure*}

An SRL system usually consists of four pipeline modules: predicate identification and disambiguation, argument identification and classification. Following most of previous work, we focus on the last two steps in standard SRL task: argument identification and classification. The predicate identification subtask is not needed in CoNLL-2009 shared task\footnote{In CoNLL-2009 task, the predicates information is already identified when testing.}, and we follow previous work \cite{he2018syntax} to handle the predicate disambiguation subtask. This work will only focus on the argument labeling subtask through sequence labeling formalization. We first describe our base model in Section \ref{basemodel}. Then we introduce the proposed associated memory network including the inter-sentence attention design and label merging strategies in Section \ref{attentionmodel}. The full model architecture is shown in Figure \ref{2}.


\subsection{Base Model \label{basemodel}}
\subsubsection*{Word Embedding}

We use the concatenation of the following embeddings as the representation for every word. (1) Random-initialized word embedding $x_i^{re}\in\mathbb{R}^{d_{re}}$ (2) GloVe \cite{pennington2014glove} word embedding $x_i^{pe}\in\mathbb{R}^{d_{pe}}$ pre-trained on 6B tokens (3)  Random-initialized part-of-speech (POS) tag embedding $x_i^{pos}\in\mathbb{R}^{d_{pos}}$ (4) Random-initialized lemma embedding $x_i^{le}\in\mathbb{R}^{d_{le}}$ (5) Contextualized word embedding derived by applying fully connected layer on ELMo embedding $x^{ce}_i \in \mathbb{R}^{d_{ce}}$ \cite{peters2018deep}, and (6) Random-initialized predicate specified flag embedding $x_i^{pred}\in\mathbb{R}^{d_{pred}}$. The final representation of each word is:
$$x_i = x_i^{re} \circ x_i^{pe} \circ x_i^{pos} \circ x_i^{le} \circ x_i^{ce} \circ x_i^{pred}$$
where $\circ$ stands for concatenation operator.

\subsubsection*{BiLSTM Encoder}
LSTM network is known to handle the dependency over long sentence well, and can effectively model the context information when encoding. Therefore, we leverage a stacked BiLSTM network $LSTM_e$ to be our encoder. It takes word embedding sequence $\mathbf{x} = [x_i]_{i=1}^{n_\mathbf{S}}$ of sentence $\mathbf{S} = [w_i]_{i=1}^{n_\mathbf{S}}$ as input ($n_\mathbf{S}$ is the length of sentence), and outputs two different hidden states $\overrightarrow{h_i}$ and $\overleftarrow{h_i}$ for word $w_i$ by processing the sequence in forward and backward directions. The final contextual representation of word $w_i$ is the concatenation of two hidden states $h_i = \overrightarrow{h_i} \circ \overleftarrow{h_i}$.

Then, we use a final softmax layer after the Bi-LSTM encoding to predict the label of each word.

\subsection{Associated Memory Network \label{attentionmodel}}
Using the base model as backbone, we introduce an associated memory network (AMN) component for further performance improvement. The proposed AMN memorizes known associated sentences and their labels, then the useful clue in the memory will be delivered to the SRL module through an inter-sentence mechanism. AMN processing includes three steps, associated sentence selection, inter-sentence attention and label merging.

\subsubsection*{Associated Sentence Selection}
We aim to utilize the associated sentences and their labels to help our model label the target sentences. For the sake of fairness, we only use the sentences in train dataset as our source. However, it is impossible to attend all the sentences in train dataset because of the extremely high computational and memory cost. Therefore, we propose a filter to select the most useful sentences from the given dataset (train dataset in this paper) when given the label-unknown sentence $\mathbf{S}$.

The filter algorithm is straightforward. First, We compute the distance of every two sentences. Then, we sort all the sentences in train dataset according to their distances with the target sentence $\mathbf{S}$, and select top $m$ sentences $\{{\bf A}_j\}_{j = 1}^m$ with the minimum distances and their label sequences $\{{\bf L}_j\}_{j = 1}^m$ as our associated attention. $m$ is the memory size. 

As for the computation of distance between two sentences, we formally consider three types of distances, which are \textit{edit distance (ED)}, \textit{word moving distance (WMD)} and \textit{smooth inverse frequency distance (SD)}, plus \textit{random distance (RD)} as baseline. These distances are defined as follows,
\begin{itemize}
	\item \textit{edit distance\label{ED}} \ \ This method uses the edit distance of the POS tag sequences of two sentences as the distance value.
	\item \textit{word moving distance} \ \ Following \cite{kusner2015word}, this method takes word moving distance of two sentences\footnote{In this paper, we use relaxed word moving distance (rwmd) for efficiency}.
	\item \textit{smooth inverse frequency distance} \ \ Following \cite{arora2016simple}, we use Euclidean distance between the SIF embedding of two sentences as the distance value. 
	\item \textit{random distance} \ \ This method returns a random value for distance computation thus lead to selecting sentences randomly in the train dataset.
\end{itemize}

\subsubsection*{Inter-sentence Attention \label{inter}}
This part aims to attain the inter-sentence attention matrix, which can be also regarded as the core memory part of the AMN. The input sentence $\mathbf{S}$ and associated sentences $\{{\bf A}_j\}_{j=1}^m$ first go through a stacked BiSLTM network ${LSTM}_a$ to encode the sentence-level information to each word representation\footnote{Here we abuse the symbol $\mathbf{S}$ and $\mathbf{A}_j$ for meaning both the word sequence $[w_i]$ and the embedded sequence $[x_i]$}:
$$\mathbf{S}' = {LSTM}_a(\mathbf{S})$$
$$\mathbf{A}_j' = {LSTM}_a(\mathbf{A}_j)\ \ j\in\{1,2,...,m\}$$
where $\mathbf{S}'=[x'_i]_{i = 1}^{n_\mathbf{S}}$ and $\mathbf{A}_j' = [x'_{j,k}]_{k=1}^{n_j}$ are the lists of new word representations, with each word representation is a vector $x' \in \mathbb{R}^{d_a}$, where $d_a$ is the size of hidden state in $LSTM_a$.

Then, for each associated sentence $\mathbf{A}'_j$, we multiply it with the input sentence representation $\mathbf{S}'$ to get the raw attention matrix $M^{raw}_{j}$.
$$M^{raw}_j = \mathbf{S}' \mathbf{A}'^T_j$$
Every element $M^{raw}_j(i,k) = x'_i \cdot x_{j,k}^{'T}$ can be regarded as an indicator of similarity between the $i^{th}$ word in input sentence $\mathbf{S}'$ and the $k^{th}$ word in associated sentence $\mathbf{A}'_j$.

Finally, we perform softmax operation on every row in $M^{raw}_j$ to normalize the value so that it can be considered as probability from input sentence $\mathbf{S}$ to associated sentence $\mathbf{A}_j$.
$$\alpha_{i,j} = f([M^{raw}_j(i,1)...,M^{raw}_j(i,n_j)])$$
$$M_j = [\alpha_{1,j},\alpha_{2,j},...,\alpha_{n_\mathbf{S},j}]$$
where $f(\cdot)$ stands for softmax function. $\alpha_{i,j}$ can be regarded as probability vector indicating the similarity between the $i^{th}$ word in sentence $\mathbf{S}$ and every word in the associated sentence $\mathbf{A}'_j$.
\subsubsection*{Label Merging}

In order to utilize the labels $\{\mathbf{L}_j\}_{j=1}^m$ of the associated sentences during decoding, a label merging needs to be done.

We use randomly initialized argument embedding $x^{ae} \in \mathbb{R}^{d_{ae}}$ to embed each argument label. Therefore, the label sequence $\mathbf{L}_j$ of associated sentence $\mathbf{A}_j$ can be written as $\mathbf{L}_j = [x^{ae}_{j,k}]_{k = 1}^{n_j}$.We treat the probability vector $\alpha_{i,j}$ as weight to sum all the elements in $L_j$ to get the associated-sentence-specified argument embedding $a_{i,j}$, which represents the attention embedding of word $w_i \in \mathbf{S}$ calculated from the $j^{th}$ associated sentence $\mathbf{A}_j$ and label $\mathbf{L}_j$.

\begin{center}
$a_{i,j} = \alpha_{i,j} \cdot \mathbf{L}_j^{T} = \sum_{k=1}^{n_j}\alpha_{i,j}(k)x^{ae}_{j,k}$
\end{center}

Because the associated sentences are different, the overall contributions of these argument embeddings should be different. We let the model itself learn how to make use of these argument embeddings. Following attention combination mechanism from \cite{libovicky2017attention}, we consider four ways to merge the label information.

\noindent\textit{1) Concatenation} All the associated argument embedding are concatenated as the final attention embeddings.

\begin{center}
$a_i = a_{i,1} \circ a_{i,2} \circ ... \circ a_{i,m}$
\end{center}

\noindent\textit{2) Average} The average value of all the associated argument embeddings is used as the final attention embedding.

\begin{center}
$a_i = \frac{1}{m}\sum_{j = 1} ^ {m} a_{i,j}$
\end{center}

\noindent\textit{3) Weighted Average\label{WA}} The weighted average of all the associated argument embedding is used as the final attention embedding. We calculate the mean value of every raw similarity matrix $M_j^{raw}$ to indicate the similarity between input sentence $\mathbf{S}$ and associated sentence $\mathbf{A}_j$, and we use the softmax function to normalize them to get a probability vector $\beta$ indicating the similarity of input sentence $\mathbf{S}$ towards all the associated sentences $\{\mathbf{A}_j\}_{j=1}^m$.

\begin{center}
$\beta = f([g(M_1^{raw}),...,g(M_m^{raw})])$
\end{center}

\noindent where $f(\cdot)$ stands for softmax function and $g(\cdot)$ represents the mean function. Then, we use the probability vector $\beta$ as weight to sum all the associated-sentence-specified attention embedding $a_{i,j}$ to get the final attention embedding $a_i$ of the $i^{th}$ word $w_i$ in input sentence $\mathbf{S}$.

\begin{center}
$a_i = \sum_{j=1}^{m}\beta(j)a_{i,j}$
\end{center}

\noindent\textit{4) Flat} This method does not use $a_{i,j}$ information. First, we concatenate all the raw similarity matrix $M_j^{raw}$ along the row.

\begin{center}
$M^{raw} = [M_1^{raw},M_2^{raw},...,M_m^{raw}]$
\end{center}

Then, we perform softmax operation on every row in $M^{raw}$ to normalize the value so that it can be considered as probability from input sentence $\mathbf{S}$ to all associated sentences ${\mathbf{A}_j}$.

\begin{center}
$\gamma_i = f([M^{raw}_{i,1},M^{raw}_{i,2}...,M^{raw}_{i,n_{all}}])$
\end{center}

\noindent where $f(\cdot)$ stands for softmax operation. $n_{all} = \sum_{j=1}^{m}n_j$ is the total length of all $m$ associated sentences.

We also concatenate the associated label information, and use $\gamma_i$ as weight to sum the concatenated label sequence as final attention embedding.

\begin{center}
$\mathbf{L} = [\mathbf{L}_1,\mathbf{L}_2,...,\mathbf{L}_j]$, \ \ $a_i = \gamma_i \cdot \mathbf{L}^T$
\end{center}

After we have the final attention embedding $a_i$, we concatenate it with word embedding $x_i$ as the input of the BiLSTM encoder $LSTM_e$.

\section{Experiments \label{exp}}
\begin{table}[t!]
	\begin{center}
		\begin{tabular}{llc}
			\hline \bf Name & \bf Meaning & \bf Value \\ \hline
			$d_{re}$ & random word embedding & 100 \\
			$d_{pe}$ & pre-trained word embedding & 100 \\
			$d_{pos}$ & POS embedding & 32 \\
			$d_{le}$ & lemma embedding & 100 \\
			$d_{ce}$ & contextualized embedding & 128 \\
			$d_{pred}$ & flag embedding & 16 \\
			$d_{ae}$ & argument embedding & 128 \\
			$m$ & memory size & 4 \\
			$k_e$ & \#$LSTM_e$ layers & 2 \\
			$k_a$ & \#$LSTM_a$ layers & 3 \\
			$d_e$ & $LSTM_e$ hidden state & 512 \\
			$d_a$ & $LSTM_a$ hidden state & 512 \\
			$r_d$ & dropout rate & 0.1 \\
			$l_r$ & learning rate & 0.001 \\
			\hline
		\end{tabular}
	\end{center}
	\caption{\label{hyper} Hyper-parameter settings (signal \#x means number of x). }
\end{table}

We conduct experiments on CoNLL-2009 \cite{hajivc2009conll} English and Chinese dataset. We use the standard training, development and test data split provided by CoNLL-2009 shared task. The word lemma, word POS are the predicted ones given in CoNLL-2009 dataset. Adam optimizer \cite{kingma2014adam} is used for training to minimize the categorical cross entropy loss. All the hyper-parameters we use are listed in Table \ref{hyper}. All parameters are learned during training, and are randomly initialized except the pre-trained GloVe \cite{pennington2014glove} word embeddings. 

For English, We independently determine the best distance calculating method and the best merging method one after another. First, we select a distance according to the results on development set and then we determine the merging method with the selected distance method. At last we explore the impact of memory size. For Chinese, we obtain the result with similar parameters as for the best model in English. The English and Chinese GloVe word embeddings are both trained on Wikipedia. The pretrained English ELMo model is from \cite{peters2018deep}, and the Chinese one is from \cite{che-EtAl:2018:K18-2}, which is hosted at \cite{fares2017word}. The model is trained for maximum 20 epochs for the nearly best model based on development set results. We re-run our model using different initialized parameters for 4 times and report the average performance\footnote{Our implementation is publicly available at \url{https://github.com/Frozenmad/AMN_SRL}.}.

\subsection{Results}
\begin{table}[t!]
	\small
	\begin{center}
		\setlength{\tabcolsep}{0.8mm}{
			\begin{tabular}{lccc}
				\hline 
				System (syntax-aware single) & P & R & F$_1$ \\ 
				\hline
				\cite{zhao2009multilingual} & - & - & 86.2 \\
				\cite{zhao2009huge} & - & - & 85.4 \\
				\cite{fitzgerald2015semantic} & - & - & 86.7 \\
				\cite{roth2016neural} & 88.1 & 85.3 & 86.7 \\
				\cite{marcheggiani2017encoding} & 89.1 & 86.8 & 88.0 \\
				\cite{he2018syntax} & 89.7 & \bf89.3 & 89.5 \\
				\bf\cite{li2018unified} & \bf90.3 & 89.3 & \bf89.8 \\
				\hline
				System (syntax-aware ensemble) & P & R & F$_1$ \\
				\hline
				\cite{fitzgerald2015semantic} & - & - & 87.7 \\
				\cite{roth2016neural} & 90.3 & 85.7 & 87.9 \\
				\bf \cite{marcheggiani2017encoding} & \bf 90.5 & \bf 87.7 & \bf 89.1 \\
				\hline
				System (syntax-agnostic single) & P & R & F$_1$ \\
				\hline
				\cite{marcheggiani2017simple} & 88.7 & 86.8 & 87.7 \\
				\cite{he2018syntax} & 89.5 & 87.9 & 88.7 \\
				\cite{cai2018full} & 89.9 & \bf 89.2 & \bf 89.6 \\
				\cite{li2018unified} & 89.5 & 87.9 & 88.7 \\
				\bf Ours ( + AMN + ELMo) & \bf 90.0 & \bf 89.2 & \bf 89.6 \\
				\hline
		\end{tabular}}
	\end{center}
	\caption{\label{englishResult} Results on CoNLL-2009 English in-domain (WSJ) test set. }
\end{table}

\begin{table}[t!]
	\small
	\begin{center}
		\setlength{\tabcolsep}{1mm}{
			\begin{tabular}{lccc}
				\hline 
				System (syntax-aware single) & P & R & F$_1$ \\ 
				\hline
				\cite{zhao2009multilingual} & - & - & 74.6 \\
				\cite{zhao2009huge} & - & - & 73.3 \\
				\cite{fitzgerald2015semantic} & - & - & 75.2 \\
				\cite{roth2016neural} & 76.9 & 73.8 & 75.3 \\
				\cite{marcheggiani2017encoding} & 78.5 & 75.9 & 77.2 \\
				\cite{he2018syntax} & \bf 81.9 & 76.9 & 79.3 \\
				\bf \cite{li2018unified} & 80.6 & \bf 79.0 & \bf 79.8 \\
				\hline
				System (syntax-aware ensemble) & P & R & F$_1$ \\
				\hline
				\cite{fitzgerald2015semantic} & - & - & 75.5 \\
				\cite{roth2016neural} & 79.7 & 73.6 & 76.5 \\
				\cite{marcheggiani2017encoding} & \bf 80.8 & \bf 77.1 & \bf 78.9 \\
				\hline
				System (syntax-agnostic single) & P & R & F$_1$ \\
				\hline
				\cite{marcheggiani2017simple} & 79.4 & 76.2 & 77.7 \\
				\cite{he2018syntax} & \bf 81.7 & 76.1 & 78.8 \\
				\cite{cai2018full} & 79.8 & 78.3 & 79.0 \\
				\bf Ours ( + AMN + ELMo) & 80.0 & \bf 79.4 & \bf 79.7 \\
				\hline
		\end{tabular}}
	\end{center}
	\caption{\label{OODtest} Results on CoNLL-2009 English out-of-domain (Brown) test set. }
\end{table}

\begin{table}[t!]
	\begin{center}
		\setlength{\tabcolsep}{0.8mm}{
			\begin{tabular}{lccc}
				\hline 
				System (syntax-aware single) & P & R & F$_1$ \\ 
				\hline
				\cite{zhao2009multilingual} & 80.4 & 75.2 & 77.7 \\
				\cite{roth2016neural} & 83.2 & 75.9 & 79.4 \\
				\cite{marcheggiani2017encoding} & 84.6 & 80.4 & 82.5 \\
				\cite{he2018syntax} &84.2 & \bf81.5 &82.8 \\
				\bf\cite{li2018unified} & \bf 84.8 & 81.2 & \bf 83.0 \\
				\hline
				System (syntax-agnostic single) & P & R & F$_1$ \\
				\hline
				\cite{marcheggiani2017simple} & 83.4 & 79.1 & 81.2 \\
				\cite{he2018syntax} & 84.5 & 79.3 & 81.8 \\
				\bf \cite{cai2018full} & 84.7 & \bf 84.0 & \bf 84.3 \\
				Ours ( + AMN + ELMo) & \bf 85.0 & 82.6 & 83.8 \\
				\hline
		\end{tabular}}
	\end{center}
	\caption{\label{chineseResult} Results on CoNLL-2009 Chinese test set. }
\end{table}

For the predicate disambiguation, we use the same one from \cite{he2018syntax} with the precisions of 95.01\% and 95.58\% on development and test sets. We compare our full model (using \textit{edit distance} and \textit{average method}) with the reported state-of-the-art models on both English and Chinese dataset. The results are in Tables \ref{englishResult}, \ref{OODtest} and \ref{chineseResult}.

For English in-domain test, our model outperforms the syntax-agnostic model in \cite{he2018syntax}, whose architecture is quite similar to our base model. Our model achieves 89.6\% in F$_1$ score, which is the same with current SOTA syntax-agnostic model \cite{cai2018full}. Besides, our result is competitive with existing syntax-aware and better than ensemble models.

The advantage is more salient on English out-of-domain test set. The F$_1$ score of our model is 79.7\%, which is 0.7\% higher than the current SOTA syntax-agnostic model \cite{cai2018full}. The result is also competitive with the best syntax-aware model \cite{li2018unified}. The comparisons show that the proposed model has a greater generalization ability.

For Chinese, starting with the similar parameters as for the best model in English, we find that attending 5 associated sentences shows a better result on Chinese. Our model achieves 83.8\% F$_1$ score, outperforming \cite{he2018syntax} with an improvement of 2.0\% in F$_1$ score. Our result is also competitive with that of \cite{cai2018full}.

Note that our method is not conflict with the one in \cite{cai2018full}, which leverages biaffine attention \cite{DBLP:journals/corr/DozatM16} for decoding. However, due to experiment cycle, we are not able to combine these two methods together. We will leave the combination as future work.


In the following part, we conduct several ablation studies on our model. All the experiments are re-run 2-4 times and the average values are reported on CoNLL-2009 English development set.

\subsection{Ablation on Distance Method}

\begin{table}[t!]
	\begin{center}
		\setlength{\tabcolsep}{0.8mm}{
		\begin{tabular}{lccc}
			\hline 
			System & P & R & F$_1$ \\ 
			\hline
			\textbf{WMD} \cite{kusner2015word} &  \bf89.1 & 87.1 & 88.1 \\
			\textbf{SD} \cite{arora2016simple} &  88.5 & \bf87.5 & 88.0 \\
			\textbf{RD} & \bf 89.1 & 87.2 & 88.1 \\
			Base Model & 88.7 & 86.9 & 87.8 \\
			\hline
			\textbf{ED} &  89.0 & \bf87.5 & \bf88.3 \\
			\hline
			\end{tabular}}
	\end{center}
	\caption{\label{distanceRes} Ablations about distance on CoNLL-2009 English development set. ED means edit distance, WMD means word moving distance, SD means SIF distance, RD means random distance.}
\end{table}

\begin{table}[t!]
	\begin{center}
		\begin{tabular}{lccc}
			\hline 
			System & P & R & F$_1$ \\ 
			\hline
			Concatenation & 88.9 & 86.6 & 87.7 \\
			\bf Average & \bf 89.0 & \bf87.5 & \bf88.3 \\
			Weighted Average & 88.7 & 87.4 & 88.1 \\
			Flat & 88.4 & 86.9 & 87.7 \\
			None & 88.7 & 86.9 & 87.8 \\
			\hline
			\end{tabular}
	\end{center}
	\caption{\label{labelRes} Ablations about label merging method on CoNLL-2009 English development set. }
\end{table}

\begin{figure}[t!]
	\centering
	\includegraphics[width=2.8in]{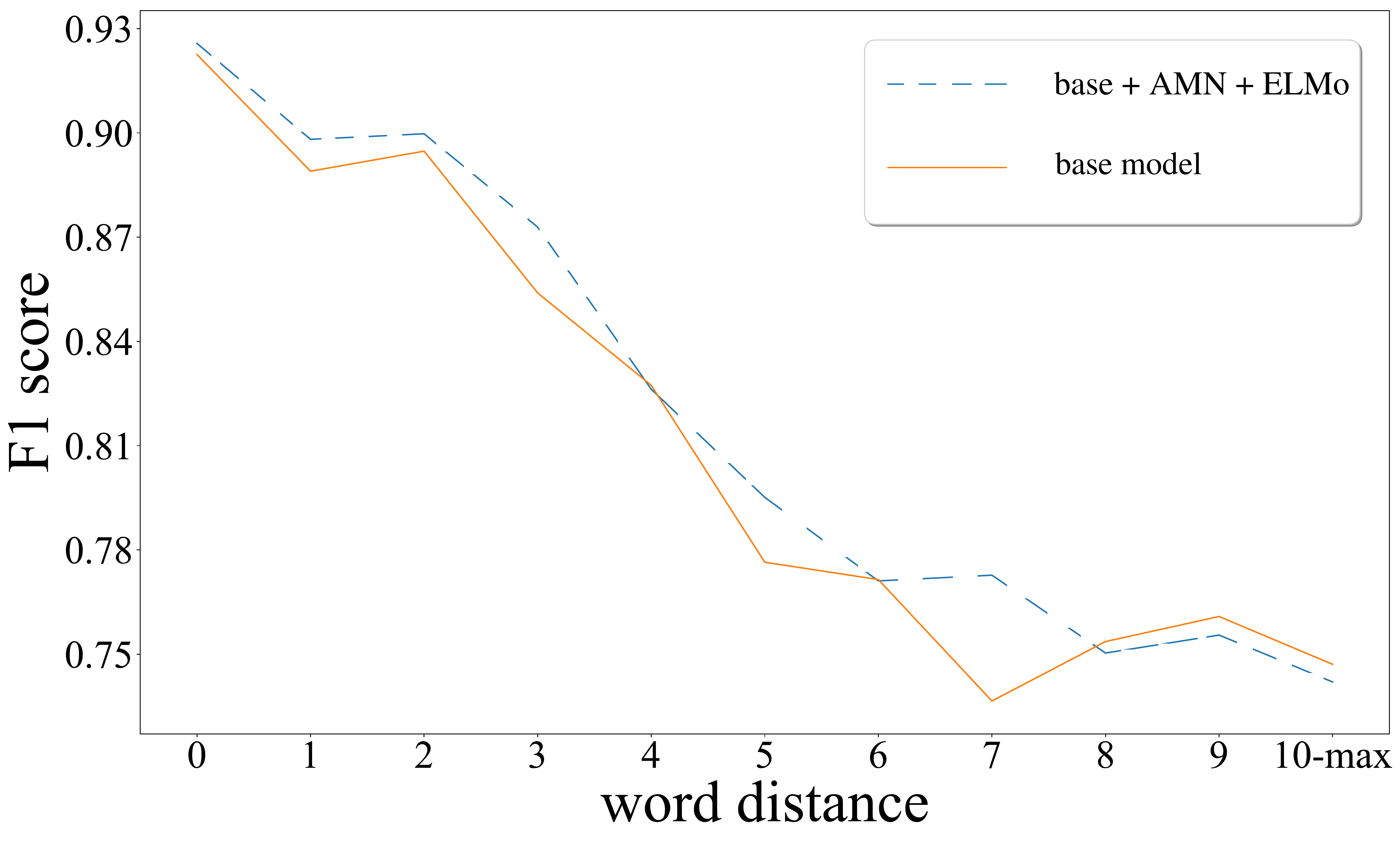}
	\caption{The comparison of our full model and base model with distance increases.\label{distance}}
\end{figure}

Table \ref{distanceRes} shows the performance of different distance calculating methods. All models use \textit{average method} for label merging, and the memory size $m$ is set to 4. It can be observed from Table \ref{distanceRes} that edit distance performs best among all the distance calculating methods, with 88.3\% F$_1$ score. All the distance calculating methods have surpassed the base model, showing that the proposed AMN is effective. Note that even the random distance model performs better than the base model, with an improvement of 0.3\% in F$_1$ score, which shows that the proposed AMN can effectively extract useful information from even poorly related sentences. Besides, associated sentence selection methods based on word embeddings like WMD and SD have similar performance with random distance (RD), which shows simple word embedding may not be good enough signal indicator to measure semantic structure similarity in SRL task. On the contrary, we may also try to explain why even the random distance selection may work to some extent. As sentences always have core arguments label such as A0, A1 and A2, associated sentences even from random selection may also have such labels, which makes them helpful to enhance SRL over these labels. This may explain why our model with randomly selected associated sentences can distinguish core arguments better.

\subsection{Ablation on Label Merging Method}

Table \ref{labelRes} shows the performance of different label merging methods. All models use \textit{edit distance} with 4 associated sentences. The result shows that \textit{Average} label merging strategy gives the best performance, achieving 88.3\% in F$_1$ score with an improvement of 0.5\% compared to the baseline model.


Note that our weighted average model does not outperform the average model, which is a surprise to us. We speculate that the current weight calculation method needs to be more improved to fit the concerned task.


\subsection{ELMo vs. AMN}

Table \ref{table:ELMo} compares the performance contribution from ELMo and AMN. 
Our model can achieve better performance only using informative clue from training set in terms of AMN design, rather than focusing on external resource like ELMo.
However, even though our baseline SRL has been enhanced by ELMo, it can still receive extra performance improvement from the propose AMN.
Note that our enhancement from the proposed AMN keeps effective when ELMo is included (a 0.5\% enhancement on baseline over the 0.3\% enhancement on ELMo baseline)

\subsection{Ablation on Memory Size}

\begin{table}[t!]
	\begin{center}
		\setlength{\tabcolsep}{0.8mm}{
			\begin{tabular}{lccc}
				\hline 
				System (syntax-aware) & P & R & F$_1$ \\ 
				\hline
				\cite{he2018syntax} & 86.8 & 85.8 & 86.3 \\
				\cite{he2018syntax} + ELMo & 87.7 & 87.0 & 87.3 \\
				\cite{li2018unified} & 87.7 & 86.7 & 87.2 \\
				\bf \cite{li2018unified} + ELMo & \bf 89.2 & \bf 87.6 & \bf 88.4 \\
				\hline
				Ours (syntax-agnostic) & P & R & F$_1$ \\
				\hline
				Base & 86.9 & 85.0 & 86.0 \\
				Base + AMN & 86.9 & 85.6 & 86.3 \\
				Base + ELMo & 88.7 & 86.9 & 87.8 \\
				\bf Ours + AMN + ELMo & \bf 89.0 & \bf 87.5 & \bf 88.3 \\
				\hline
		\end{tabular}}
	\end{center}
	\caption{\label{table:ELMo} AMN vs. ELMo, the performance comparison on English development set.}
\end{table}

\begin{figure}[t!]
	\centering
	\includegraphics[width=0.48\textwidth]{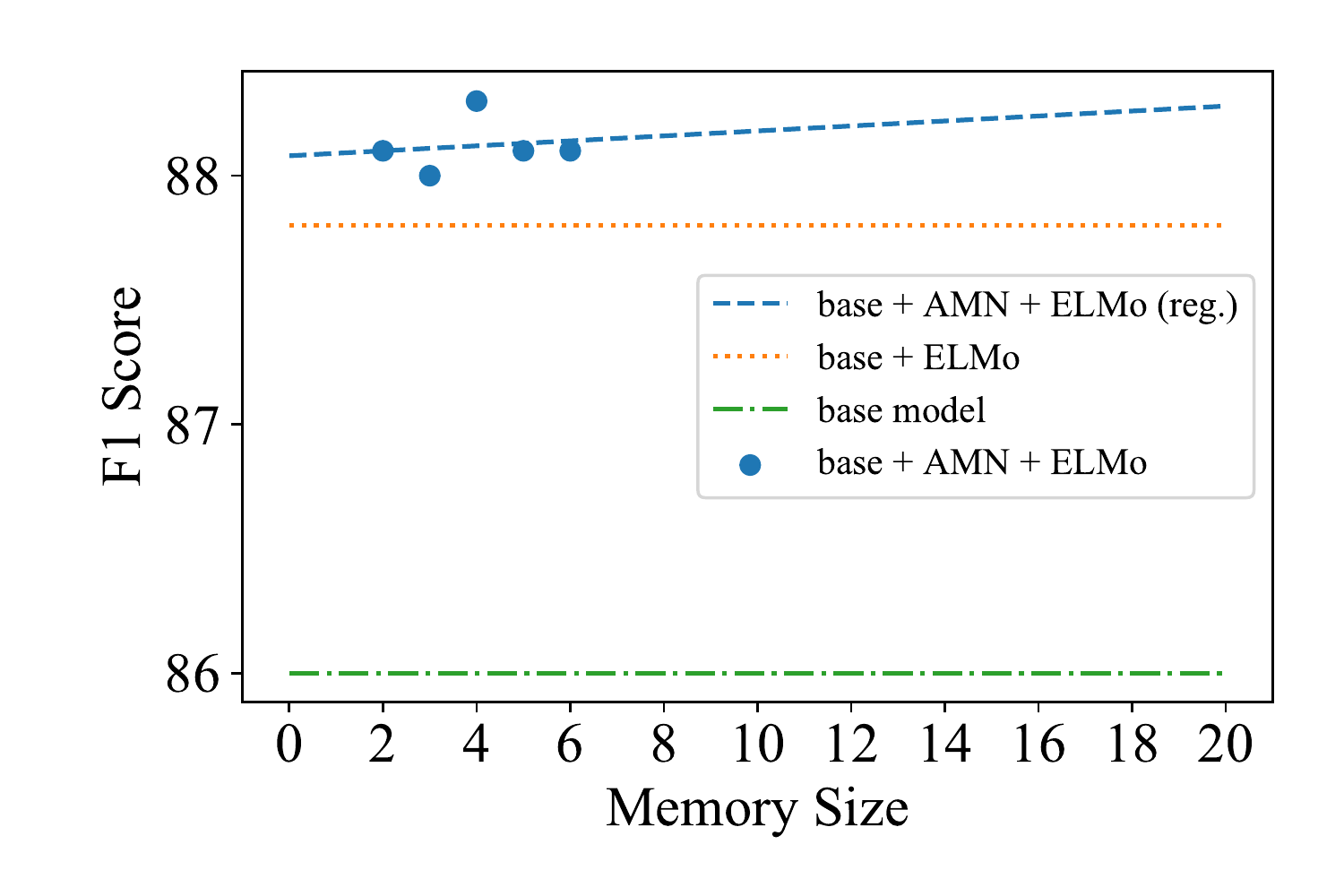}
	\caption{Model performance on English development set with different memory sizes, in which \textit{base}+\textit{AMN}+\textit{ELMo} (\textit{reg}.) indicates the general trend of our base model enhanced by the AMN when the memory size is enlarged.\label{a2}}
\end{figure}

We show the effect of different memory size in Figure \ref{a2}. Note that more associated sentences means more cost on time and space. We test memory size $m$ from 2 to 6 (which reaches the limit under experiment setting in 11G GPU). We also fit the measured points with a linear function (the blue line in Figure \ref{a2}). The performance of our model has a general trend of increasing when the memory size becomes larger, which shows the potential of the proposed AMN.


\begin{figure*}[htbp]
	\begin{minipage}[t]{0.48\textwidth}
	  \centering
	  \includegraphics[width=2.9in]{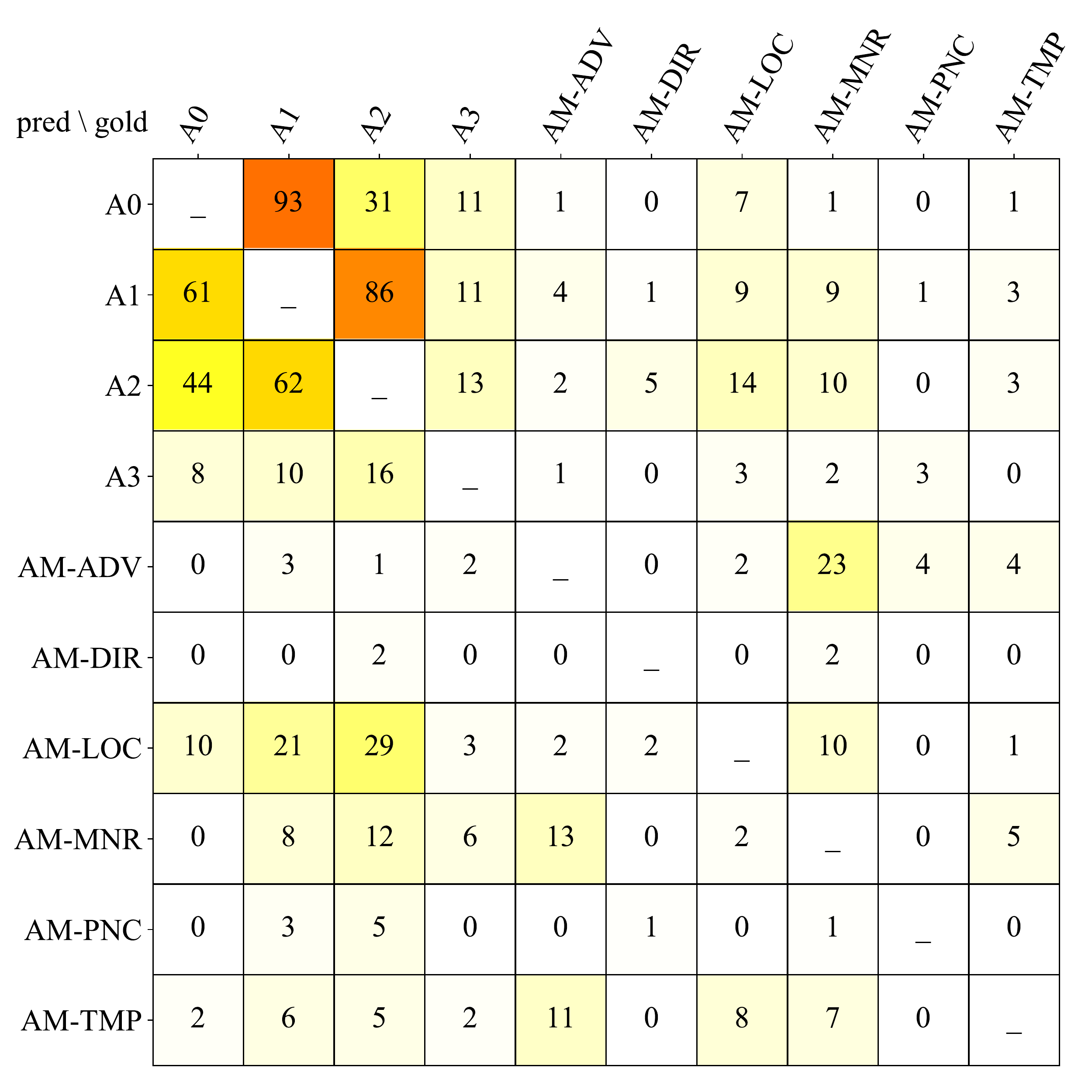}
	  \caption{Confusion matrix for labeling errors in base model.\label{hotpoint1}}	
	\end{minipage}
	\begin{minipage}[t]{0.48\textwidth}
	  \centering
	  \includegraphics[width=2.9in]{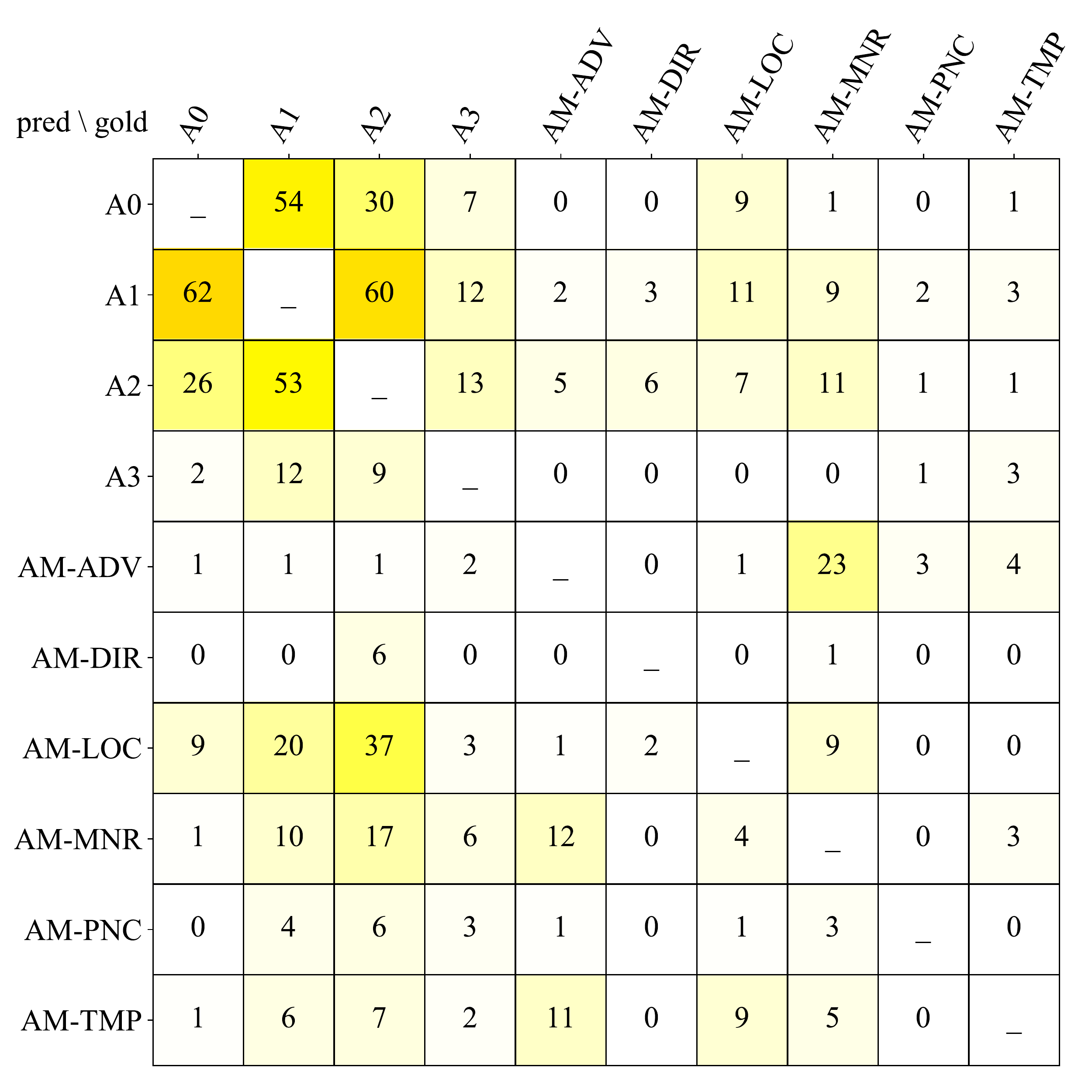}
	  \caption{Confusion matrix for labeling errors in proposed model.\label{hotpoint2}}
	\end{minipage}      
  \end{figure*}

\subsection{Analysis on Confusion Matrix}

To further understand the advance of the proposed method, we conduct an error type break down analysis. Figures \ref{hotpoint1} and \ref{hotpoint2} show the confusion matrices of labeling errors in the baseline model and our model on development set, respectively. We only show the main and most informative type of arguments. Every number in these figures stands for the times of occurrence. Comparing these two confusion matrixes shows that the proposed model makes fewer mistakes between core arguments such as $A0$, $A1$, and $A2$. AMN indeed helps when labeling them. It is also noted that, as in \cite{he2017deep, Tan2017Deep}, the model still easily confuses ARG2 with AM-DIR, AM-LOC and AM-MNR.

\subsection{Analysis of Performance on Distance}

We compare the performance concerning with the distance of argument and predicate on our best model and base model in Figure \ref{distance}, from which we can observe that our model performs better nearly at any distance.

\subsection{Case Study on AMN}
\begin{figure}[t!]
	\centering
	\includegraphics[width=2.8in]{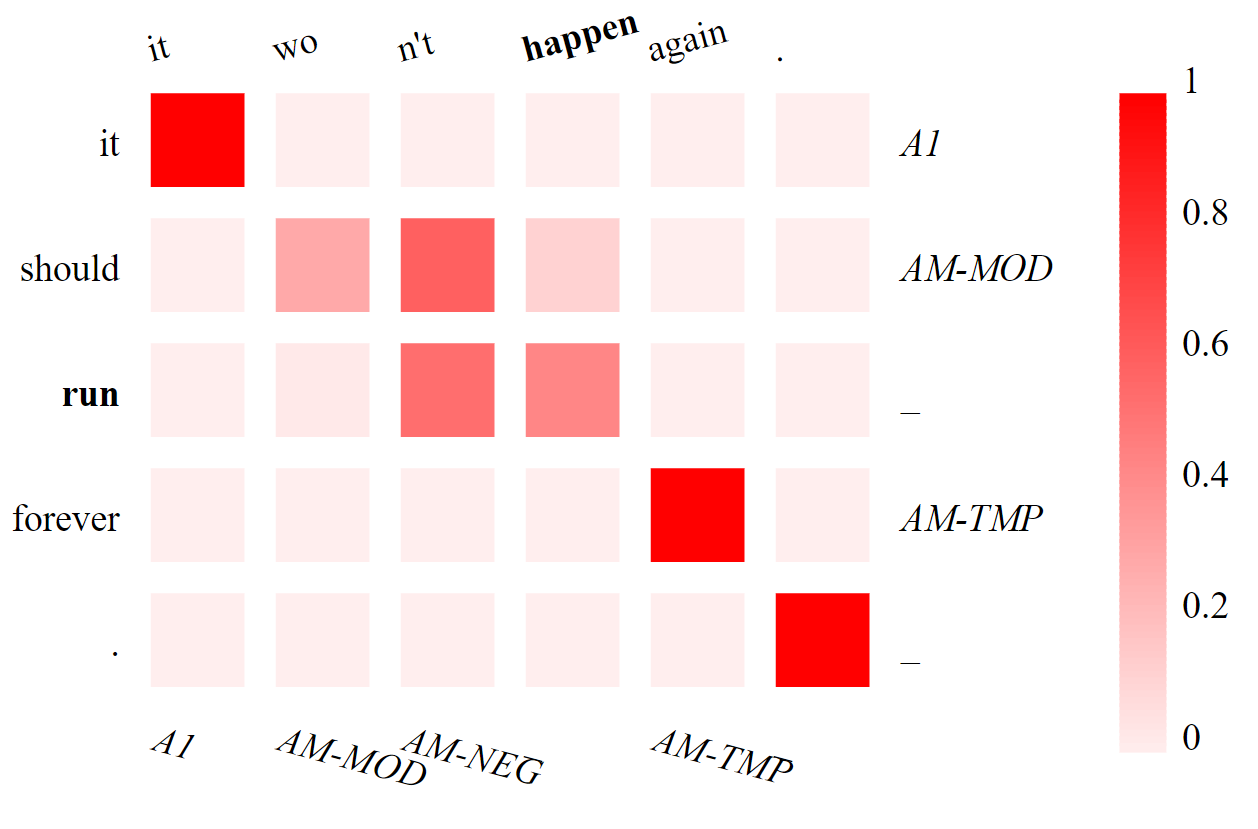}
	\caption{Visualization of similar matrix $M$. The input sentence is at the left of the matrix, with its golden argument label at the right. The associated sentence is at the top of the matrix, with its golden argument label at the bottom. Their predicate is bolded.\label{simi}}
\end{figure}

To explore how the AMN works in the model, we visualize the similarity matrix $M$ of some sentences from development set in Figure \ref{simi}. The input sentence is

\textbf{it}$_{\scriptsize A1}$ \textbf{should}$_{\scriptsize AM-MOD}$ \textbf{run}$_{\scriptsize v}$ \textbf{forever}$_{\scriptsize AM-TMP}$.

\noindent
And the associated sentence is

\textbf{it}$_{\scriptsize A1}$ \textbf{wo}$_{\scriptsize AM-MOD}$ \textbf{n\'t}$_{\scriptsize AM-NEG}$ \textbf{happen}$_{\scriptsize v}$

\textbf{again}$_{\scriptsize AM-TMP}$.

\noindent
The current predicates are \textbf{run}, \textbf{happen} respectively. The visualization shows that inter-sentence attention can find and align the word in the similar context correctly, which shows that the proposed AMN is reasonable and effective.



\section{Related Works \label{related}}
Early attempts \cite{pradhan2005semantic,zhao2009multilingual,zhao2009semantic,zhao2013integrative,roth2014composition} to the SRL task were mainly linear classifiers. The main focus was how to find proper feature templates that can best describe the sentences. \cite{pradhan2005semantic} utilized a SVM classifier with rich syntactic features. \cite{toutanova2008global} took the structural constraint into consideration by using a global reranker. \cite{zhao2009huge} adopted a maximum entropy model with large scale feature template selection. \cite{roth2014composition} explored the distributional word representations as new feature to gain more powerful models.

Recently, a great attention has been paid on neural networks. \cite{zhou2015end} proposed an end-to-end model using stacked BiLSTM network combined with CRF decoder without any syntactic input. \cite{marcheggiani2017simple} explored the predicate-specified encoding and decoding and also provided a syntax-agnostic LSTM model. \cite{he2017deep} followed \cite{zhou2015end} and analyzed all popular methods for initialization and regularization in LSTM network.

By considering that our approach also borrows power from the memory, the proposed inter-sentence attention in our AMN shares features with memory networks, which was proposed in \cite{Weston2015Memory} with motivation that memory may reduce the long-term forgetting issues.
\cite{sukhbaatar2015end} and  \cite{miller2016key} later further improved this work. However, we use quite different mechanisms to store the memory, and the effectiveness of our model needs a carefully designed attention mechanism to handle the sequence-level information distilling.

Attention mechanism was first used by  \cite{bahdanau2014neural} in machine translation. Recently, \cite{Tan2017Deep} and \cite{strubell2018linguistically} proposed to use self-attention mechanism in SRL task. \cite{cai2018full} leveraged the biaffine attention \cite{DBLP:journals/corr/DozatM16} for better decoding performance. Different from all the existing work, we instead introduce an inter-sentence attention to further enhance the current state-of-the-art SRL.

\section{Conclusions and Future Work \label{future}}
This paper presents a new alternative improvement on strong SRL baselines. We leverage memory network which seeks power from known data, the associated sentences, and thus is called associated memory network (AMN). The performance of our model on CoNLL-2009 benchmarks shows that the proposed AMN is effective on SRL task.

As to our best knowledge, this is the first attempt to use memory network in SRL task. There is still a large space to explore along this research line. For example, our weighted average method may need more carefully improved. Our model can be built over the biaffine attention which has been verified effective in \cite{cai2018full}\footnote{As this paper is submitting, we get to know the work \cite{li2018dependency}, which has taken both strengths of biaffine and ELMo. We leave the verification of our proposed method over this new strong baseline in the future.}, and the encoder in our model can be improved with more advanced forms such as Transformer \cite{vaswani2017attention}. At last, as this work is done on a basis of quite limited computational resources, only one piece of nVidia 1080Ti (11G graphic memory), much plentiful available computational resource will greatly enable us to explore more big model setting (i.e., larger memory size $m$) for more hopefully better performance improvement.


\bibliography{naaclhlt2019}
\bibliographystyle{acl_natbib}

\end{document}